\documentclass[letterpaper, 10 pt, conference]{ieeeconf}  

\IEEEoverridecommandlockouts                              

\usepackage{epsfig} 
\usepackage{epstopdf}
\usepackage{mathptmx} 
\usepackage{graphicx}
\usepackage{subfigure}
\usepackage{tabularx,capt-of}
\usepackage{hhline}
\usepackage{cite}
\usepackage{amssymb,amsmath,amsfonts}
\usepackage{algorithm}
\usepackage[noend]{algpseudocode}
\usepackage{thmtools}
\usepackage{todonotes}
\usepackage{url}
\usepackage{comment}
\usepackage{tabularx}
\usepackage{chngcntr}
\usepackage{multirow}
\usepackage{cancel}
\usepackage{multicol}
\newtheorem{definition}{\textbf{Definition}}

\newtheorem{problem}{\textbf{Problem}}

\DeclareMathOperator*{\argmax}{argmax}

\newcommand{\ie}[1]{\textit{i.e.,}}

\makeatletter
\def\BState{\State\hskip-\ALG@thistlm}

\makeatother

\title{\LARGE \bf
Efficiently Identifying Hotspots in a Spatially Varying Field with Multiple Robots
}

\author{Varun Suryan$^{1}$ and Pratap Tokekar$^{2}$
\thanks{$^{1}$Varun Suryan. Department of Computer Science, University of Maryland College Park, USA.
        {\tt\small varun.suryan.01@gmail.com}}%
\thanks{$^{2}$Pratap Tokekar. Department of Computer Science, University of Maryland,
        Collge Park, USA.
        {\tt\small tokekar@umd.edu}}%
}

\begin{document}

\maketitle
\thispagestyle{empty}
\pagestyle{empty}

\begin{abstract}
In this paper, we present algorithms to identify environmental hotspots using mobile sensors. We examine two approaches: one involving a single robot and another using multiple robots coordinated through a decentralized robot system. We introduce an adaptive algorithm that does not require precise knowledge of Gaussian Processes (GPs) hyperparameters, making the modeling process more flexible. The robots operate for a pre-defined time in the environment. The multi-robot system uses Voronoi partitioning to divide tasks and a Monte Carlo Tree Search for optimal path planning. Our tests on synthetic and a real-world dataset of Chlorophyll density from a Pacific Ocean sub-region suggest that accurate estimation of GP hyperparameters may not be essential for hotspot detection, potentially simplifying environmental monitoring tasks.
\end{abstract}

\section{INTRODUCTION}
\begin{figure}[ht]
  \centering
  \includegraphics[width=0.6\linewidth]{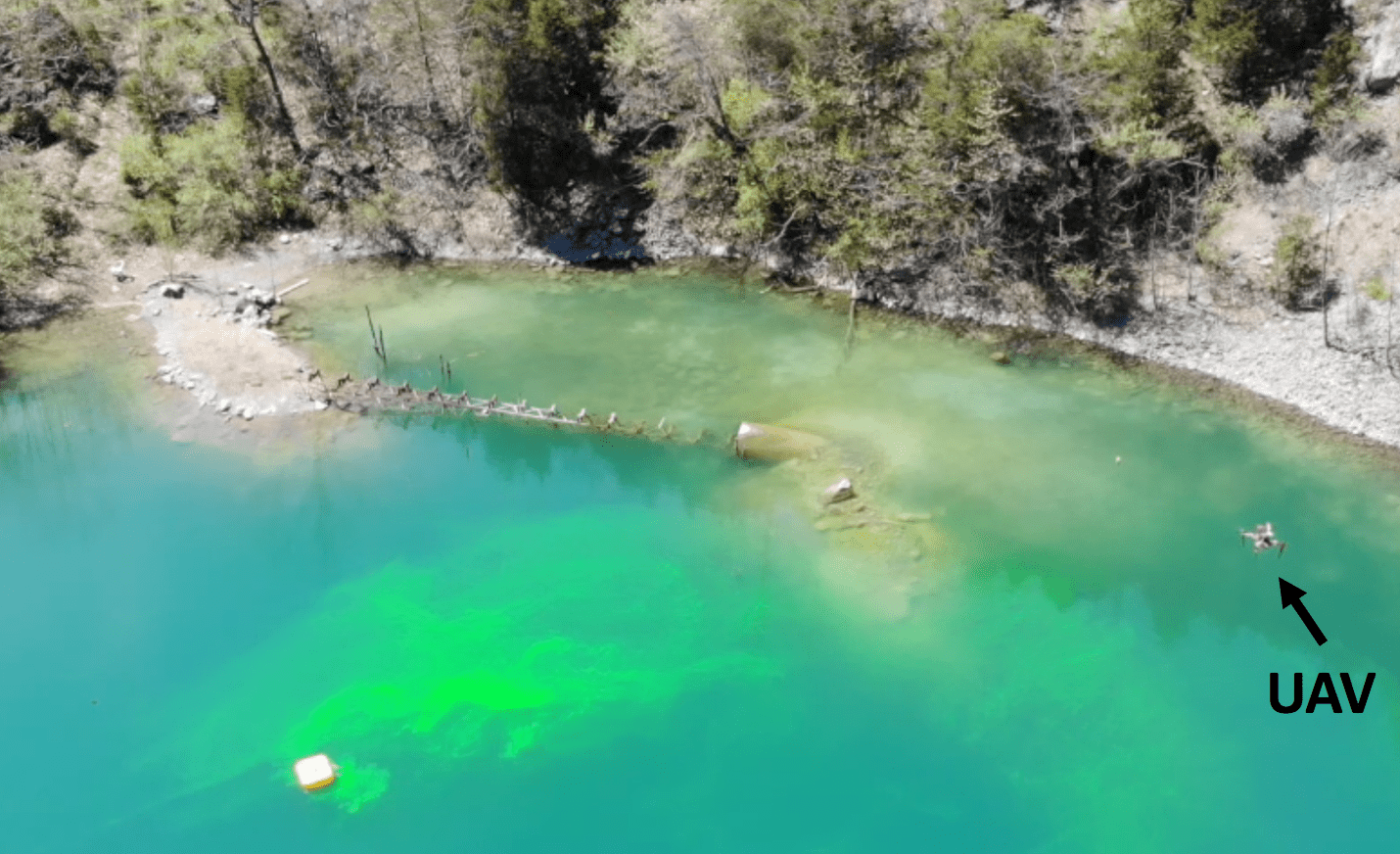}
  \caption{An unmanned aerial vehicle (UAV) flying over a lake to find the chemical spill hotspots~\cite{sung2018online}.\label{fig:smart_bridge}}
\end{figure}

Mobile robots are increasingly used in collecting information in multitudes of scenarios. For example, a farmer can send a robot to collect the measurements of organic matter in different sub-regions of the farm~\cite{tokekar2016sensor, suryan2020learning} to understand the soil chemistry~\cite{aktar2009impact}. Robots can detect any environmental anomalies, such as a chemical spill in a water body which can have a significant impact on marine life~\cite{blanchard2020informative}. An aerial robot (Figure~\ref{fig:smart_bridge}) can be used to monitor relatively large areas at once~\cite{sung2018online}. In another application, robots can be deployed in a nuclear power plant to monitor potential leakages by measuring radiation levels~\cite{moore1985usa}. By identifying the sites of higher nuclear radiation using robot sensors, we can efficiently find any potential leakage. In these scenarios, one would be better off just by identifying the hotspot (i.e., maxima) instead of learning the entire environment accurately like in our prior work~\cite{suryan2020learning}.

Our goal is to plan the paths to identify the hotspots with a single as well as multiple mobile robots. For a single robot, we present a Monte Carlo Tree Search (MCTS)-based~\cite{Browne2012} planning algorithm that uses an Upper Confidence Bound (UCB)-style~\cite{srinivas2009gaussian} exploration and works with or without the knowledge of true Gaussian Processes (GP) hyperparameters. In general, GP hyperparameters are optimized during the process and can be a computationally prohibitive task. For the multi-robot case, we present a dynamic partitioning scheme that splits the environment amongst the robots such that no robot is required to cover an especially large portion of the environment. However, instead of partitioning the environment just based on the size, we use the GP estimates and the size of the environment to determine the partitions. Specifically, our partitioning is based on Voronoi tessellation~\cite{tanemura1983new} and the UCB metric~\cite{inproceedingsKobilarov, srinivas2009gaussian}. This partitioning scheme can work with several planners and find hotspots efficiently. We also allow for the robots to operate in a decentralized fashion with periodic connectivity for coordination. 

\section{Related Work}
The hotspot identification issue aligns with problems like source-seeking in Informative Path Planning literature~\cite{7526683, DBLP:journals/corr/abs-1809-10611}. Chen and Liu introduced Pareto MCTS, an anytime multi-objective planning method addressing exploration vs. exploitation~\cite{Liu-RSS-19}. While many informative planning studies assume known hyperparameters~\cite{10.5555/1625275.1625631, krause2008near, binney2013optimizing, suryan2018sensor, suryan2020learning}, online planning estimates them during execution. Binney et al.~\cite{binney2013optimizing} used initial run data for estimation. Kemna et al.~\cite{kemna2018pilot} utilized pilot surveys for hyperparameter initialization, accounting for their time in overall planning. MCTS has been commonly used in informative path planning and hotspot identification~\cite{Liu-RSS-19, Browne2012, xiao2022nonmyopic}. They have been shown to have consistencies in balancing the exploration-exploitation efficiently in many applications~\cite{gelly2012grand, browne2012survey}. Our algorithm AdaptGP-MCTS uses GP-UCB values as the reward heuristics and balances the exploration-exploitation. The performance of UCB planners has been shown to be sensitive with respect to $\beta$ value~\cite{10.5555/3020751.3020809}. In this work, we use a squared root growth of $\beta$ which has been proved to achieve better performance on terminal regret~\cite{tolpin2012mcts}.

Multi-Robot Systems (MRS) have been actively deployed in precision agriculture~\cite{barrientos2011aerial, kazmi2011adaptive}, and environmental monitoring and exploration~\cite{dhariwal2004bacterium, dunbabin2012robots, ouimet2013collective}. One of the major challenges in MRS is dividing the task between robots efficiently, especially in practical scenarios when the robots operate in a decentralized manner~\cite{https://doi.org/10.48550/arxiv.2203.02865}. Voronoi partitioning is a common approach for multi-robot coordination used in various domains, such as exploration and mapping with ground vehicles, including spatial partitioning~\cite{1013690, doi:10.1177/0278364913515309, 5650551,6385730, 6798666}. Kemna et al. used a dynamic Voronoi partitioning approach based on the entropy in a decentralized fashion~\cite{7989245}. They repeatedly calculate weighted Voronoi partitions for the space. Each vehicle then runs informative adaptive sampling within its partition. The vehicles can share information periodically. Wenhao et al. presented an adaptive sampling algorithm for learning the density function in multi-robot sensor coverage problems using a Mixture of GP models~\cite{8460473}.

\section{Problem Formulation}
We assume that the spatial field under consideration defined over a 2-dimensional environment $U\in\mathbb{R}^2$ is an instance of a GP, $F$. $F$ is defined by a covariance function of the form,
\begin{equation}\label{eq:kerenldef}
    C_Z(x, x') = \sigma^2\exp\left(-\frac{(x-x')^2}{2l^2}\right); \forall x, x'\in U,
\end{equation}
defined by a squared-exponential kernel and the hyperparameters $\sigma^2$ and $l$ are not known.

\begin{problem}[Terminal Regret]\label{ch3prob:terminal}
Given an operating time budget $T$, plan a trajectory under budget $T$ for a mobile robot that obtains measurements from $U$, and reports the location of maxima of the spatial field $f$ at the end, \ie,
\begin{equation*}
\begin{aligned}
& \underset{}{\text{minimize}}
& & f(x^*) - f(\hat{x}), \\
& \text{subject to}
& & len(\tau) + n\eta \leq T.
\end{aligned}
\end{equation*}
$\tau$ denotes the tour of the robot. The robot travels at unit speed, obtains one measurement in $\eta$ units of time, and collects $n$ total measurements. $\hat{x}$ is the location of the maxima of the predicted field while $x^*$ is the location of the maxima of the true spatial field. We do not know $x^*$ and we also do not know $f$. We only know the GP prediction $\hat{f}$. The task is to use $\hat{f}$ to be able to predict $x^*$.
\end{problem}

\begin{problem}[Multi-robot Hotspot ID]\label{prob:terminal}
Given an operating time budget $T$, plan a set of trajectories under budget $T$ for a set of $k$ mobile robots that obtain measurements from the environment $U$, and report the location of maxima of the spatial field $f$ at the end, \ie,
\begin{equation*}
\begin{aligned}
& \underset{}{\text{minimize}}
& & f(x^*) - f(\hat{x}), \\
& \text{subject to}
& & \underset{i\in\lbrace1,\ldots,k\rbrace}{\text{max}} len(\tau_i)  + n_i\eta \leq T.
\end{aligned}
\end{equation*}
$\tau_i$ denotes the tour of the $i^{th}$ robot. Robots travel with unit speed and obtain one measurement in $\eta$ units of time. Here, let $i^{th}$ robot collect $n_i$ total measurements.
\end{problem}

\section{Algorithms}\label{ch3sec:alg}
We start with the algorithm for a single robot followed by the multi-robot version.
\subsection{Single Robot}
AdaptGP-MCTS (Algorithm~\ref{alg:adaptAlg}) shows the main function that calls the planner MCTS shown in Line~\ref{line:call_planner}. Once the planner gives the next measurement location, the robot goes there and collects the measurement. AdaptGP-MCTS monotonically decreases the length scale and monotonically increases the signal variance so that the GP model can capture more complex function candidates~\cite{Berkenkamp2019}. Eliminating the need to optimize hyperparameters at each step by using AdaptGP-MCTS alleviates the cubic complexity of the hyperparameter optimization. AdaptGP-MCTS starts with an initial $\sigma_0$ and $l_0$ of the GP hyperparameters. The new updated values of hyperparameters are used to get the mean and variance estimate in the next iteration in Line~\ref{line:get_new_mu}. In Line~\ref{line:coll_meas}, we collect the measurement at location $x_t$. This measurement is perturbed by the sensor noise $\epsilon$ modeled as a standard normal distribution with mean zero mean and $\omega^2$ variance. $\omega^2$ is assumed to be known \emph{a priori}. Once the operating budget is exhausted, we do a full GP hyperparameter optimization (Line~\ref{line:final_update}). Finally, the location of the predicted maxima is reported (Line~\ref{line:report_loc}) where the posterior mean attains its maximum value.

\begin{algorithm}[htp]
\caption{AdaptGP-MCTS}\label{alg:adaptAlg}
\begin{algorithmic}[1]
\BState \textbf{Input:} Initial hyperparameters $\sigma_0=1$ and $\mathbf{l_0}=diam(Env)$, $\mathbf{X}$ = \{\}, $\mathbf{y}$ = \{\}, Planner().
\BState \textbf{while} $t \leq~\textnormal{Total time budget}~T$
\State  \ \ \ \ $\hat{\mu}_{t}(x), \hat{\sigma}_{t}(x) \gets GP.Predict(\mathbf{X}, \mathbf{y})$\label{line:get_new_mu}
\State  \ \ \ \ $x_t~\gets~Planner(\hat{\mu}_{t}(x), \hat{\sigma}_{t}(x), t)$ \label{line:call_planner}
\State \ \ \ \ $y_t = f(x_t) + \epsilon$\label{line:coll_meas}
\State \ \ \ \ $\mathbf{X}.append(x_t); \mathbf{y}.append(y_t)$ \label{line:append}
\State \ \ \ \ Update $\sigma_t = \sigma_0\log(t); \mathbf{l_t} = \mathbf{l_0}/\log(t)$\label{line:change_hyper}
\BState Do a full GP hyperparameter optimization with $(\mathbf{X}, \mathbf{y})$ \label{line:final_update}
\BState Estimate the posterior mean $\hat{\mu}$
\BState  return  $\argmax_{x\in U} \hat{\mu}(x)$\label{line:report_loc}
\end{algorithmic}
\end{algorithm}

Now we discuss the planner which is based on the idea of MCTS and uses GP-UCB values as the reward heuristics. The pseudocode for the planner is given in the Algorithm~\ref{alg:adaptMCTS}.
\begin{algorithm}[htp]
\caption{GP-MCTS}\label{alg:adaptMCTS}
\begin{algorithmic}[1]
\BState \textbf{Input:} $\hat{\mu}_{t}(x), \hat{\sigma}_{t}(x), t$ .
\BState \textbf{while} $within~budget$
\State  \ \  $v^\prime \gets\textnormal{Selection}(v)$ 
\State \ \   $v_{new} \gets\textnormal{Expansion}(v^\prime)$ \label{line:expansion}
\State \ \   $r_\mu +\beta^{1/2} r_\sigma \gets\textnormal{Simulation}(v_{new})$ \label{line:define_reward}
\State \ \   $\textnormal{Backpropagation}\left(v_{new}, r_\mu +\beta^{1/2} r_\sigma\right)$
\BState \textbf{end procedure}
\BState \textbf{function} Selection($v$)
\State \ \ \ while $v$ is fully expanded
\State \ \ \ \ \ \ \ $v\gets\argmax_{child\in v.{children}} \frac{Q(child)}{n_{child}}+ 2\sqrt{\frac{\log(n_v)}{n_{child}}}$ \label{line:pareto}
\State \ \ \  return $v$
\end{algorithmic}
\end{algorithm}
In the Backpropagation step, we use the GP-UCB values to update the values for ancestral nodes. For reward calculation, we use a root squared growth of $\beta^{1/2}$ in terms of the number of measurements collected:
\begin{enumerate}
    \item{Mean:} To encourage the exploitation, \ie, $r_\mu = \hat{\mu}_t(x)$,
    \item{Variance:} To encourage the exploration, \ie, $r_\sigma = \hat{\sigma}_t(x)$.
\end{enumerate}

\subsection{Multiple Robots}
Our multi-robot algorithm uses Voronoi regions for dynamic partitioning after each epoch.
\begin{definition}
Given a set of points $p_1, p_2, \ldots, p_n$ in the plane S, a Voronoi diagram divides the plane S into $n$ Voronoi regions with the following properties~\cite{tanemura1983new}:
\begin{itemize}
    \item Each point $p_i$ lies in exactly one region.
    \item If a point $q \in S$ lies in the same region as $p_i$, then the Euclidian distance from $p_i$ to $q$ will be shorter than the Euclidean distance from $p_j$ to $q$, where $p_j$ is any other point in S.
\end{itemize}
\end{definition}
The points $p_1, \ldots, p_n$ are called generator points for the Voronoi partitions. We use UCB values defined in~\cite{srinivas2009gaussian} (the denominator in Equation~\ref{eq:ucbExp}) as the weights from our GP model to estimate the weighted centroids of a Voronoi cell. Let $(x_1^1, x_2^1), \ldots, (x_1^m, x_2^m)_i$ be the set of $m$ points in $i^{th}$ Voronoi partition. Then its centroid can be calculated as follows,

\begin{equation}
\begin{split}
        Centroid(Vor_i) &= \\
    \sum_{k=1}^{k=m}&\frac{{(x_1^k, x_2^k)_i(\hat{\mu}_{t}(x_1^k, x_2^k)_i + \beta_t\hat{\sigma}_{t}(x_1^k, x_2^k)_i)}}{\hat{\mu}_{t}(x_1^k, x_2^k)_i + \beta_t\hat{\sigma}_{t}(x_1^k, x_2^k)_i}.
\end{split}
    \label{eq:ucbExp}
\end{equation}
Here, $\hat{\mu}_{t}(x_1^k, x_2^k)_i$, and $\hat{\sigma}_{t}(x_1^k, x_2^k)_i$ are the GP mean and variance at location $(x_1^k, x_2^k)_i$ respectively, and $\beta_t$ is the parameter that controls the exploration-exploitation.



\begin{algorithm*}[!htp]
\caption{Voronoi-TrueGP-MCTS}\label{alg:Voronoi}
\begin{algorithmic}[1]
\Procedure{}{}
\BState \textbf{Input:} Hyperparameters $\sigma_0$ and $l_0$, Planner().
\BState for epoch = [1:n]
\State \hspace{.3cm} Create $k$ individual copies $GP1, GP2, \ldots, GP_k$ of the combined GP model $GP_{combined}$
\State \hspace{.3cm} Calculate Voronoi regions using robot locations as the point generators \label{line:calc_voronoi} 
\State \hspace{.3cm}    for t = [1:m]:
\State \hspace{1.3cm}   Plan and execute the next robot step computed using Planner() in their respective Voronoi region.
\State \hspace{1.3cm}   Collect the next environment samples, $(x_1^t, x_2^t)_1, \ldots, (x_1^t, x_2^t)_k$
\State \hspace{1.3cm}   Update the GP models: $GP1.update(x_1^t, x_2^t)_1, \ GP2.update(x_1^t, x_2^t)_2 \ldots,\ GPk.update(x_1^t, y_2^t)_k$
    
\State \hspace{.3cm}    Combine the samples from all robots and update the combined GP model using all the samples.

\BState \textbf{end procedure}
\EndProcedure 
\end{algorithmic}
\end{algorithm*}

In Algorithm~\ref{alg:Voronoi}, robots operate for $n$ epochs and take $m$ steps per epoch. They begin from set start points. Voronoi regions for these robots are derived using their current positions. During an epoch, each robot's path is mapped out using the Planner() function within its specific Voronoi area. Within an epoch, robots cannot exchange information. Thus, the Planner() function relies solely on the data each robot individually knows during that epoch. Measurements taken are identified by robot number; e.g., $(x_1^t, x_2^t)_1, (x_1^t, x_2^t)_2$ represent data collected by Robots 1 and 2 at time $t$ respectively. Once the epoch concludes, robots share data, and we update the collective GP model, $GP_{combined}$, with all cumulatively gathered measurements. Voronoi partitions are then recalculated with current robot positions (Line~\ref{line:calc_voronoi}). If the GP hyperparameters are not known, the AdaptGP-MCTS planner for a single robot, detailed in Algorithm~\ref{alg:adaptAlg}, can be applied.

\section{Empirical Evaluation}\label{sec:experiment}
We start by presenting our empirical results with the case where the GP hyperparameters are assumed to be known. We call this strategy TrueGP-MCTS.

Our tests use Chlorophyll density data from a Pacific Ocean square sub-region, covering longitude from -155.5 to -129.5 and latitude from 9.0 to 35. We modeled an environment using these coordinates, studying a synthetic spatial field. Locations within are treated as search tree nodes. Robots at any location have five motion primitives, uniformly distributed in the $[-\frac{\pi}{4}, \frac{\pi}{4}]$ range, acting as current node children. The MCTS build iteration cap is 50. We used a random policy for roll-outs, back-propagating average GP-UCB values as rewards. Roll-outs don't have fixed simulation steps. Instead, they're based on the remaining time budget minus the node's depth from the root. This approach promotes more exploration early on but diminishes as the mission progresses and the environment becomes familiar~\cite{Browne2012}.

\begin{figure}[htp]
	\centering
	\includegraphics[width=0.7\columnwidth]{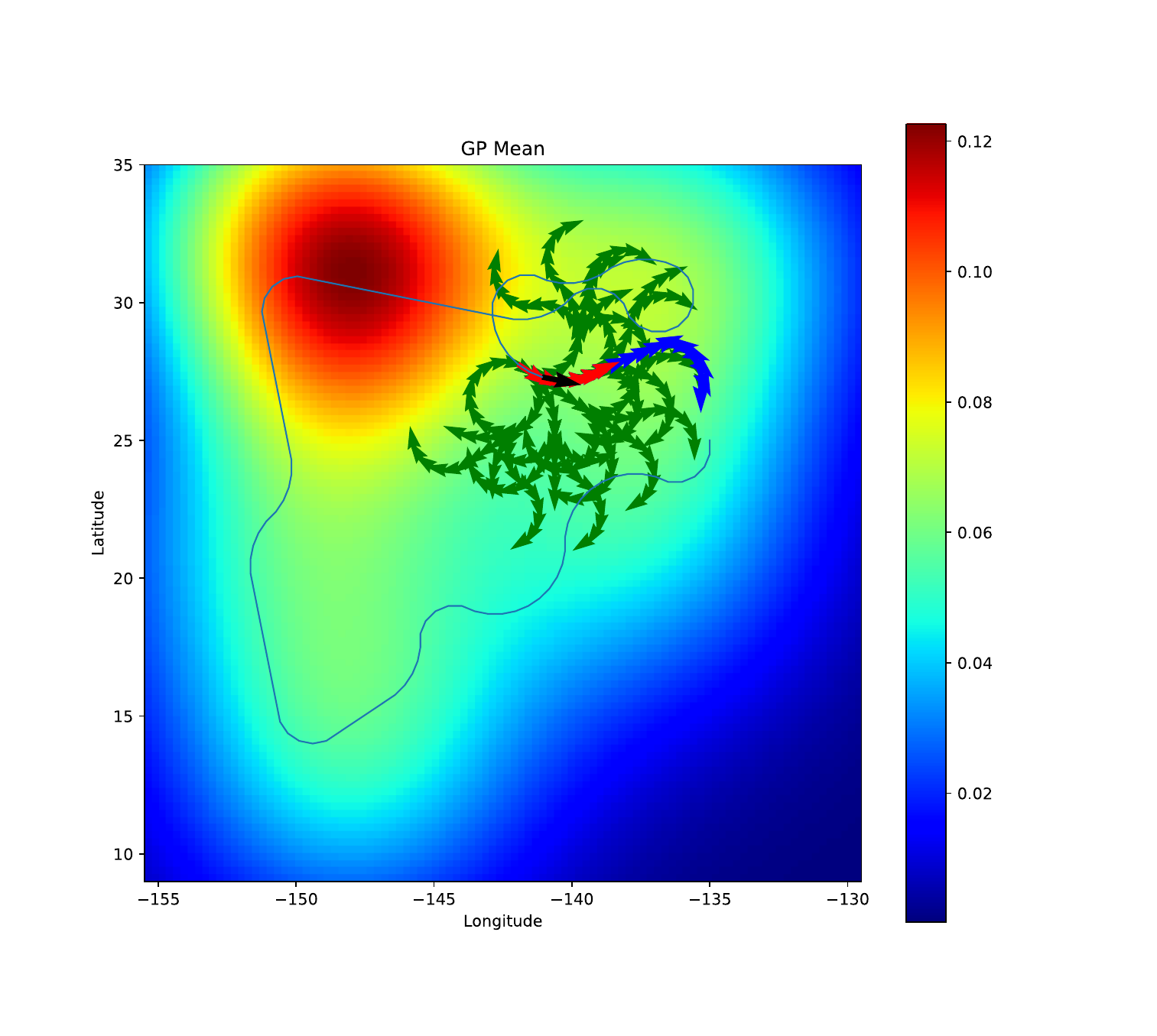}
	\caption {The robot has five motion primitives.}
   \label{fig:mcts_demo}
\end{figure}

An instance of an MCTS tree for a robot is shown in Figure~\ref{fig:mcts_demo}. The green arrows represent the entire tree and the blue arrows represent the best trajectory based on this built tree. The blue path shows the robot path until that moment in time and the background heatmap represents the learned GP mean by the robot of the underlying spatial field. For the Expansion Step in Algorithm~\ref{alg:adaptMCTS} (Line~\ref{line:expansion}), we expand randomly on any of the unvisited children.

\subsection{Synthetic Field}
\begin{figure}[htp]
	\centering
	\includegraphics[width=0.7\columnwidth]{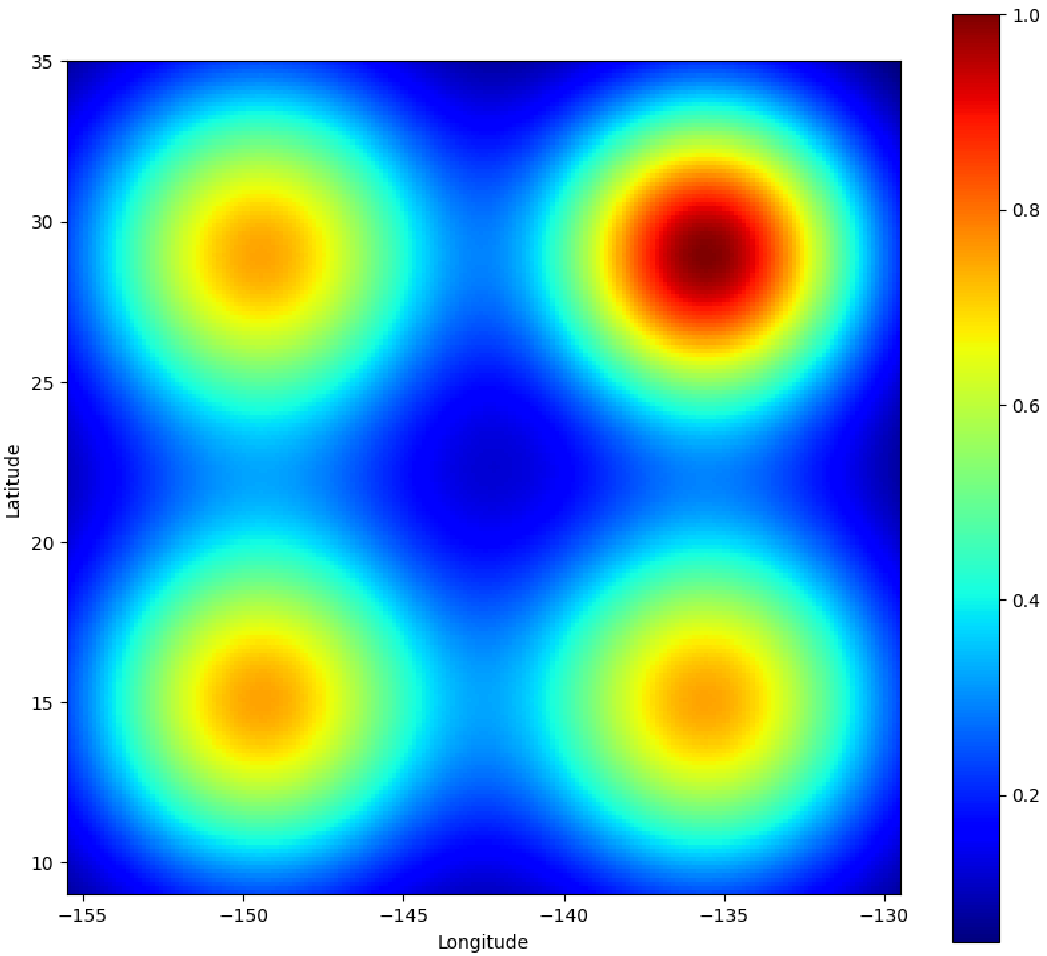}
	\caption {The environment has four locations of maxima, three of which are local maxima.}
   \label{fig:fourmaxima}
\end{figure}

We construct a complex spatial field (Figure~\ref{fig:fourmaxima}) that has four locations of maxima, three of which are local maxima. For our experiments, we start the robot near the lower left corner from (-149.0, 16.0) so as to trick it into collecting measurements and spending time near one of the local maxima. The actual hotspot is located near the top right corner at (-135.6, 29) where the field attains a maximum value of 1 and a minimum value of 0.
We estimated the hyperparameters \emph{apriori} using a $30\times30$ grid on this field and minimizing the negative log marginal likelihood of the values at those grid locations. The GP squared-exponential hyperparameters $\sigma_0, l1, l2, \omega^2$ for this field were estimated to be $0.251, 5.04, 5.04, 10^{-5}$ respectively.

The sensor noise standard deviation was set to 0.05 (5\% of the spatial field range). The robot plans the path using an MCTS planner with GP-UCB values as the node rewards where the GP variance was multiplied with the square root of $2\sqrt{t}\log\left(\frac{|D|\pi^2}{6\delta}\right)$ as $\beta_t$ (termed as GP-UCBE in the plots). Here, $|D|$ denotes the number of grid locations used for estimating the GP mean and variance. We used a grid of resolution $130\times130$. Hence, $|D|$ is $16900$ in our case and we choose $\delta$ to be equal to 0.1~\cite{srinivas2009gaussian}. We run ten missions for the robot that starts from (-149.0, 16.0). We compare the performance of the TrueGP-MCTS planner with a Boustrophedon (BST) path.

Table~\ref{table:fouthotspot} shows the average mission Percent Terminal Regret, Percent Average Cumulative Regret, Percent Root Mean Squared Error (RMSE) all with respect to the range of the spatial field (\ie, 1), Percent Distance with respect to the diagonal of the environment. The TrueGP-MCTS outperforms the BST on all metrics. The BST exhibits a higher standard deviation in its performance, influenced by the orientation of its pattern, which might occasionally lead to quick hotspot detection or prolonged searches. In contrast, TrueGP-MCTS maintains a more consistent, uniform exploration of the environment.

\begin{figure}[htp]
	\centering
	\includegraphics[width=0.8\columnwidth]{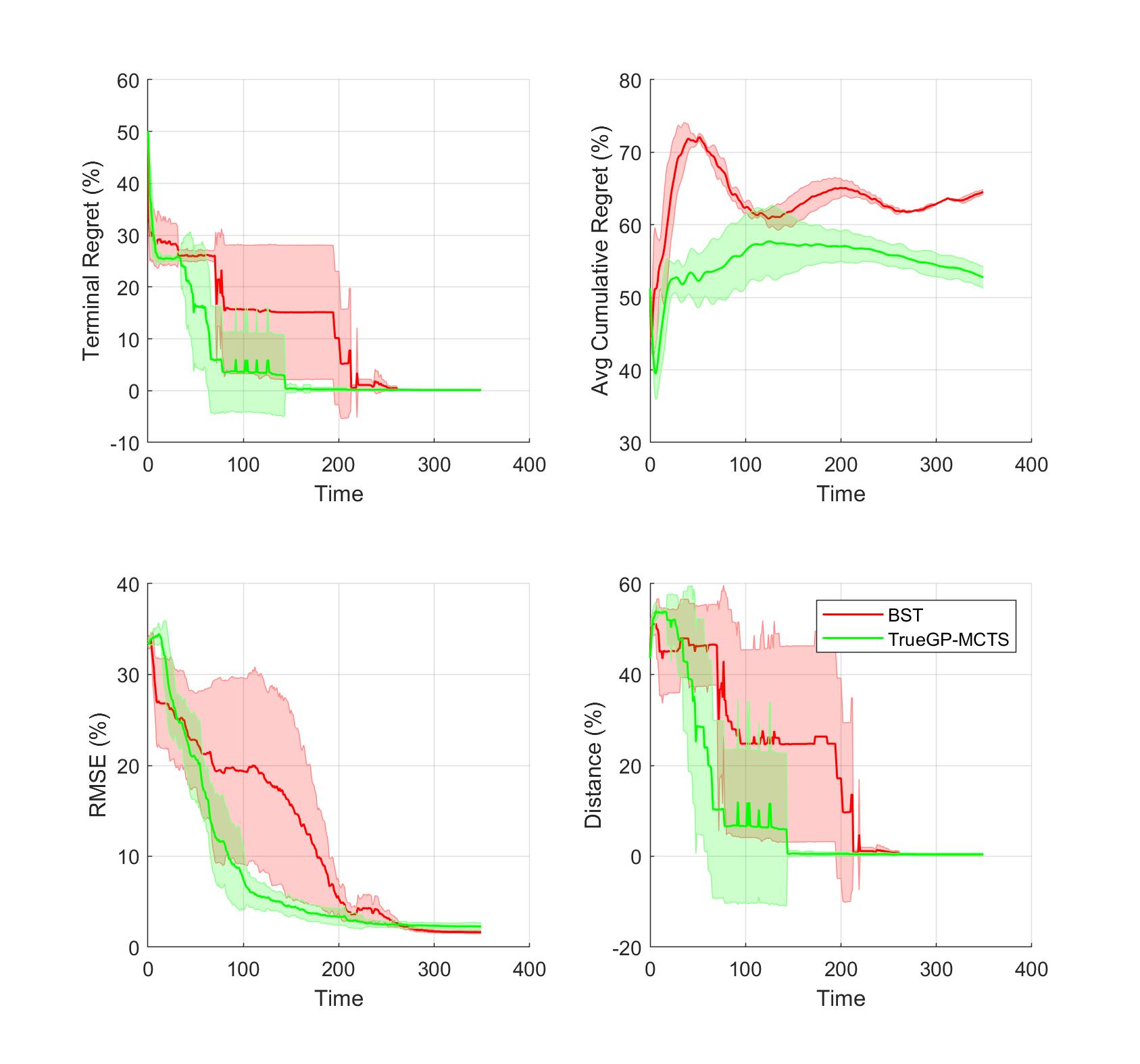}
	\caption {The sensor noise standard deviation was set to 5\%.}
   \label{fig:hot1rstd5}
\end{figure}

 \begin{table}
\centering
\caption{\textnormal{The time budget is 350 units with the first subcolumn displaying the BST pattern and the second showing TrueGP-MCTS.}}
\begin{tabular}{|c|c|c|}
\hline
& BST & TrueGP-MCTS\\
 \hline
Terminal Regret& $11.7130   \pm 4.8586$ & $5.3964\pm    2.1146$ \\
Avg Cumulative Regret & $63.6402\pm    0.7974 $ & $54.8814   \pm 1.9327$\\
RMSE& $ 11.9767   \pm 4.2813 $ & $8.2699  \pm  1.0573 $\\
Distance& $ 19.7206   \pm 9.3801$ & $9.7927\pm    4.8182 $ \\
\hline
\end{tabular}
\label{table:fouthotspot}
\end{table}
Figure~\ref{fig:hot1rstd5} shows the same metrics as Table~\ref{table:fouthotspot}. We can see that in the beginning, BST and TrueGP-MCTS have almost the same performance in terms of terminal regret and distance. However, with a medium budget, the TrueGP-MCTS explores the environment efficiently and converges quickly to report the hotspot location.

\subsection{Chlorophyll Dataset}
\begin{figure}
	\centering 
    \subfigure[Geographical Subregion\label{fig:chlorogeog}]{\includegraphics[width=0.4\columnwidth]{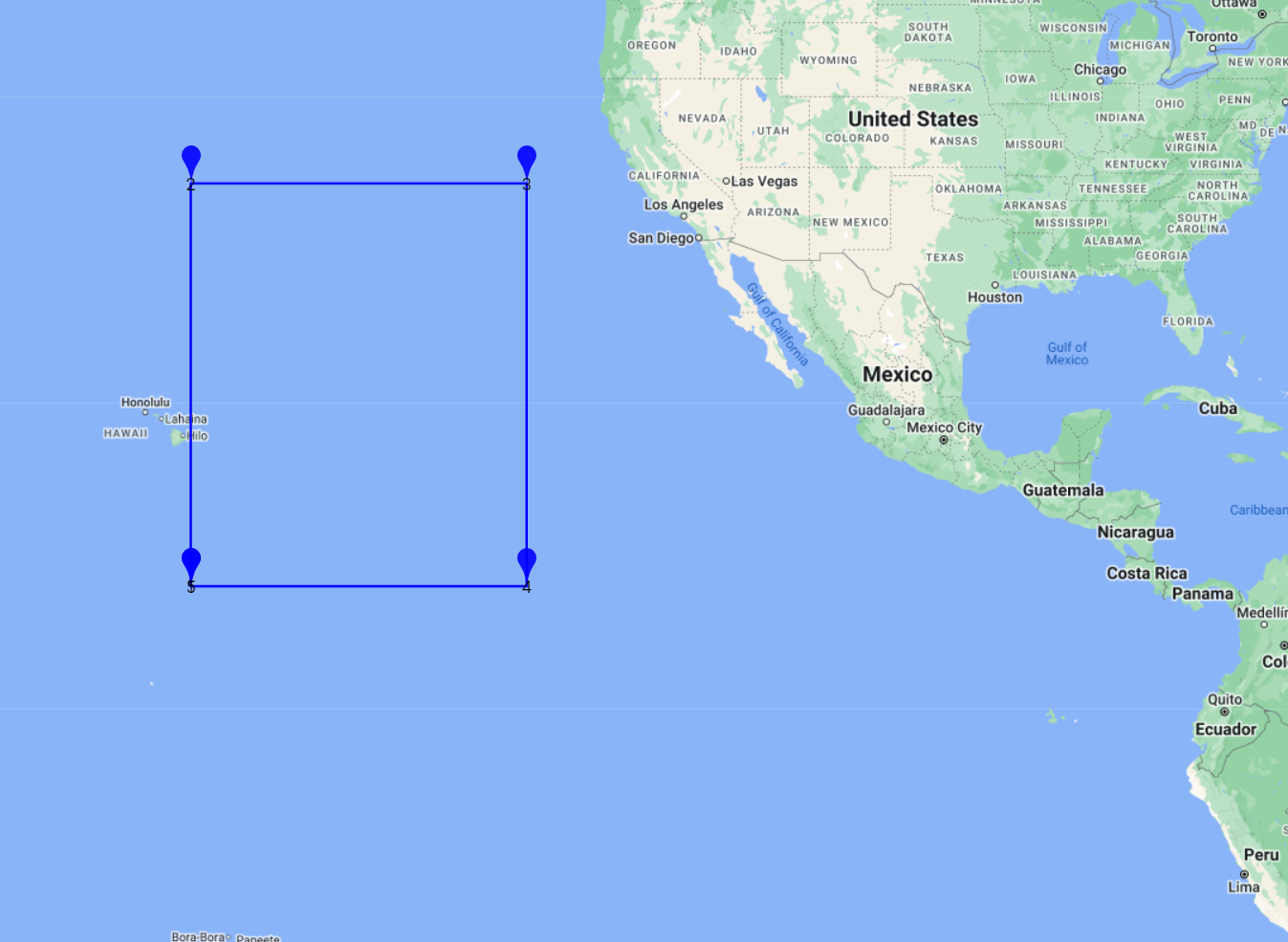}}
    \subfigure[Chlorophyll Concentration ($mg/m^3$)\label{fig:chlorodensity}]{\includegraphics[width=0.55\columnwidth]{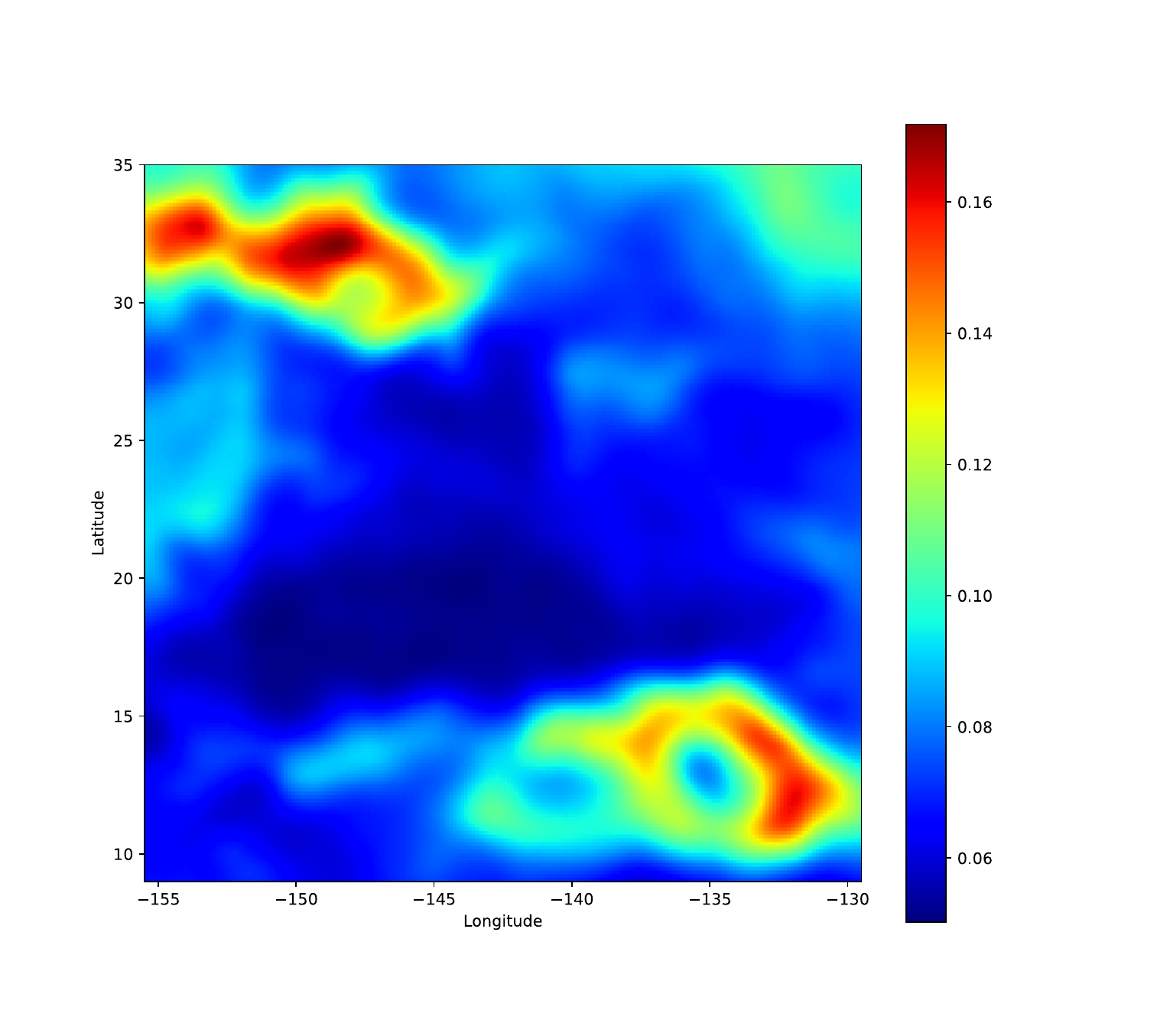}}
	\caption{The environment has longitude expansion from -155.5 to -129.5 and latitude expansion from 9 to 35.}
   \label{fig:env_chlorophyll}
\end{figure}

We evaluate the performance of our algorithms on a real-world dataset of Chlorophyll concentration measured on Oct 8, 2021, obtained from NASA Earth Observations from a Pacific Ocean subregion shown in Figure~\ref{fig:chlorogeog}. The actual Chlorophyll concentration ($mg/m^3$) is shown in Figure~\ref{fig:chlorodensity}. The data collected is from a square region spanning the geographical coordinates, longitude expansion from -155.5 to -129.5, and latitude expansion from 9 to 35 (Figure~\ref{fig:chlorogeog}) at 0.5 degree geo-coordinate grid resolution. To query a value at any non-grid location, we used a radial basis function for interpolating and assumed that the interpolated values were the true values at that non-grid location.

The hotspot is located at (-148.67, 32.11) where the Chlorophyll density attains the maximum value equal to 0.17 $mg/m^3$ and the lowest density value is 0.05 $mg/m^3$. We estimated the hyperparameters \emph{apriori} using a $30\times30$ grid on this field and minimizing the negative log marginal likelihood of the values at those grid locations. The GP squared-exponential hyperparameters $\sigma_0, l1, l2, \omega^2$ for this field were estimated to be $0.0483, 2.33, 1.99, 10^{-5}$ respectively.
The sensor values are simulated as a normal distribution with the mean as the actual value at the measurement location. The sensor noise standard deviation was set to 0.006 (5\% of the spatial field range).

We run ten missions for the robot that starts from (-142, 18). This starting location was chosen closer to the local maxima and is more likely to trick the robot from identifying the actual hotspot. We compare the performance of the TrueGP-MCTS planner with a Boustrophedon (BST) path. Table~\ref{ch4table:1} shows all the metrics similar to Table~\ref{table:fouthotspot} for the Chlorophyll dataset. The TrueGP-MCTS planner outperforms Boustrophedon on Terminal Regret, RMSE, and Distance and comparably on Cumulative Regret. The TrueGP-MCTS outperforms the BST path and keeps accumulating cumulative regret by continuously exploring the environment even though it has already found the hotspot. Hence, while it might not be always traveling in the high-value regions (resulting in a higher cumulative regret) its GP-Mean estimate still has the maxima aligned with the actual hotspot location.
 
 \begin{table}
\centering
\caption{\textnormal{The time budget is 350 units.}}
\begin{tabular}{|c|c|c|}
 \hline
 &  BST & TrueGP-MCTS  \\
 \hline
Terminal Regret&  $32.21\pm 16.38$ & $26.37\pm 7.76$\\
Avg Cumulative Regret  & $76.52 \pm 1.01$ & $68.1\pm 1.91$\\
RMSE& $26.59\pm 5.16 $ & $20.58\pm 1.08$ \\
Distance & $34.54\pm 12.14$ & $16.84\pm 8.57 $\\
 \hline
 \end{tabular}
\label{ch4table:1}
\end{table}

\begin{figure}[htp]
	\centering
	\includegraphics[width=0.7\columnwidth]{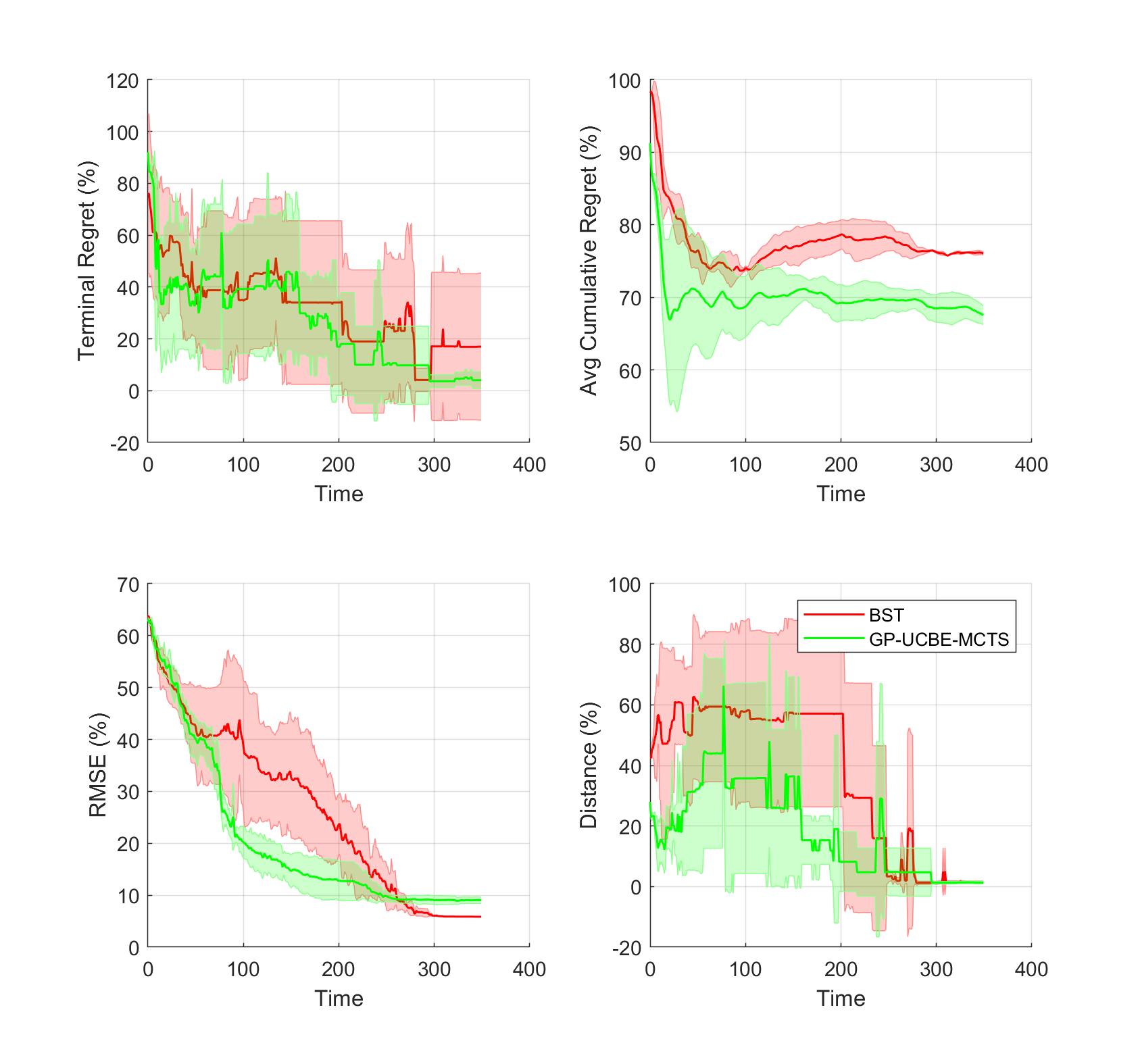}
	\caption {The sensor noise standard deviation was set to 5\%.}
   \label{fig:1rstd5}
\end{figure}

Figure~\ref{fig:1rstd5} shows the same metrics as Table~\ref{table:fouthotspot}. In the beginning, BST and TrueGP-MCTS have almost the same performance in terms of terminal regret and distance. However, with a medium budget, the TrueGP-MCTS explores the environment efficiently and converges quickly to report the hotspot location.

\subsection{Unknown GP Hyperparameters}
\begin{figure}[htp]
	\centering
	\includegraphics[width=0.6\columnwidth]{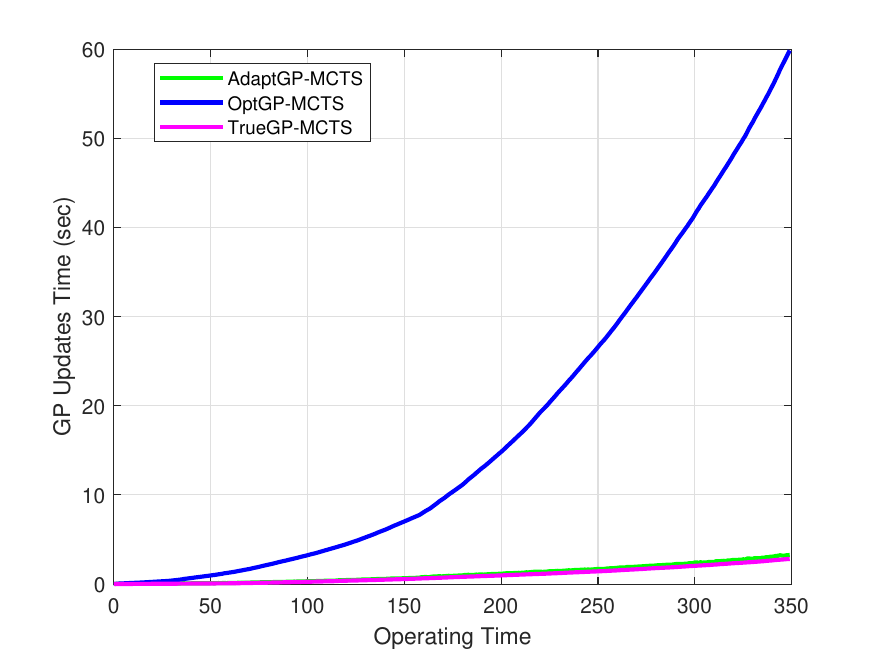}
	\caption {The sensor noise standard deviation was set to 5\%.}
   \label{fig:1rstd5time}
\end{figure}

We now present the AdaptGP-MCTS and compare its performance with TrueGP-MCST (known hyperparameters) and OptGP-MCTS (optimized at every timestep). We did the experiments with a single robot on the synthetic spatial field. Table~\ref{table:adaptvstrue} displays metrics akin to Table~\ref{table:fouthotspot}. Over ten missions, TrueGP-MCTS initially performs better than the rest, with a marginal lead over OptGP-MCTS. OptGP-MCTS shows notable initial variability, likely due to its hyperparameters being path-dependent, causing variations across missions. Hence, on a low operating time budget and unknown hyperparameters, one can use OptGP-MCTS. Figure~\ref{fig:1rstd5time} shows the cumulative GP operations time versus the operating budget. TrueGP-MCTS and AdaptGP-MCTS have almost the same computation time but OptGP-MCTS complexity increases significantly. However, as the robot spends more time in the environment, the AdaptGP-MCTS catches up and the performance difference diminishes to less than $3\%$.

\begin{table}
\centering
\caption{\textnormal{The top, middle, and bottom sub-tables display metrics for the time budgets 1-100, 101-200, and 201-350, respectively.}}
\begin{tabular}{|c|c|c|}
 \hline
 $58.02\pm13.10$  & $17.58\pm8.30$ & $17.15\pm3.31$ \\
 $59.27\pm4.62$ & $53.50\pm2.80$ & $50.70\pm2.46$  \\
 $34.33\pm4.67$ & $22.19\pm4.98$ & $21.06\pm3.73$\\
 $48.38\pm4.13$ & $27.64\pm16.46$ & $28.88\pm5.21$\\
 \hline
 \hline
 $18.13\pm 14.51$ & $2.33\pm3.34$ & $0.56\pm0.66$\\
 $62.58\pm 0.96$ & $59.71\pm3.12$ & $54.25\pm3.85$\\
 $8.92\pm 1.00$ & $5.98\pm1.09$ & $4.43\pm0.47$ \\
 $21.23\pm 20.68$ & $4.38\pm 7.20$ & $1.10\pm1.26$\\
 \hline
 \hline
 $3.13\pm2.07$ & $0.07\pm0.10$ & $0.17\pm0.22$ \\
 $62.39\pm1.14$ & $57.19\pm1.61$  & $53.88\pm1.11$\\
 $6.36\pm0.29$ & $2.62\pm0.38$ & $2.71\pm0.24$\\
 $3.28\pm2.84$ & $0.30\pm 0.21$ & $0.44\pm0.34$\\
\hline
\end{tabular}
\label{table:adaptvstrue}
\end{table}

\subsection{Multiple Robots}
We conducted tests using four robots on the synthetic spatial field depicted in Figure~\ref{fig:fourmaxima}. The robots start near the bottom left, enticing them to gather measurements near one of the peaks. We compared three algorithms in each scenario:

\begin{enumerate}
    \item{Boustrophedon (BST):} Every robot individually follows a boustrophedon pattern.
    \item{No partition: Robots can explore the entire environment anytime without being restricted to their Voronoi partition.}
    \item{Site partition: Robots are limited to their Voronoi partitions, determined by their last surfacing event.}
\end{enumerate}

We compare the Voronoi partitioning and No partitioning in terms of the time taken by them to find all the hotspots (4 for the synthetic field). Table~\ref{ch4table:hot4rtimetaken} shows the earliest time for four robots to detect 1, 2, 3, and 4 hotspots. The Voronoi partitioning achieves better exploration and outperforms No Partitioning when it comes to finding multiple hotspots. Table~\ref{ch4table:hot3rtimetaken} shows the earliest time for three robots to detect 1, 2, 3, and 4 hotspots.

\begin{table}
\centering
\caption{\textnormal{Averaged over 15 trials for four robots and the noise was set to 5\%.}}\vspace{0.1in}
\begin{tabular}{|c|c|c|}
\hline
& No Partition&Voronoi\\
 \hline
1 Hotspot&  $3.70 \pm  1.15$ & $ 4.60  \pm 1.77$ \\
2 Hotspots& $25.40 \pm  8.09$ & $11.40\pm    5.22$\\
3 Hotspots& $37.40 \pm  6.96$ & $21.10\pm    5.70$\\
2 Hotspots& $55.70 \pm  14.10$ & $34.20\pm    9.35$\\
 \hline
\end{tabular}
\label{ch4table:hot4rtimetaken}
\end{table}

\begin{table}
\centering
\caption{\textnormal{Averaged over 15 trials for three robots and the noise was set to 5\%.}}
\begin{tabular}{|c|c|c|}
\hline
& No Partition & Voronoi\\
 \hline
1 Hotspot&  $4.73 \pm  2.15$ & $ 4.86  \pm 1.88$ \\
2 Hotspots& $31.46 \pm  5.73$ & $11.93\pm    4.00$\\
3 Hotspots& $50.06 \pm  12.69$ & $20.53\pm    8.81$\\
4 Hotspots& $71.13 \pm  15.00$ & $44.33\pm    9.80$\\
 \hline
\end{tabular}
\label{ch4table:hot3rtimetaken}
\end{table}

\subsection{Chlorophyll Dataset}
\begin{figure}
	\centering 
    \subfigure[No partitioning path after the robots have spent 50 units of time.]{\includegraphics[width=0.9\columnwidth]{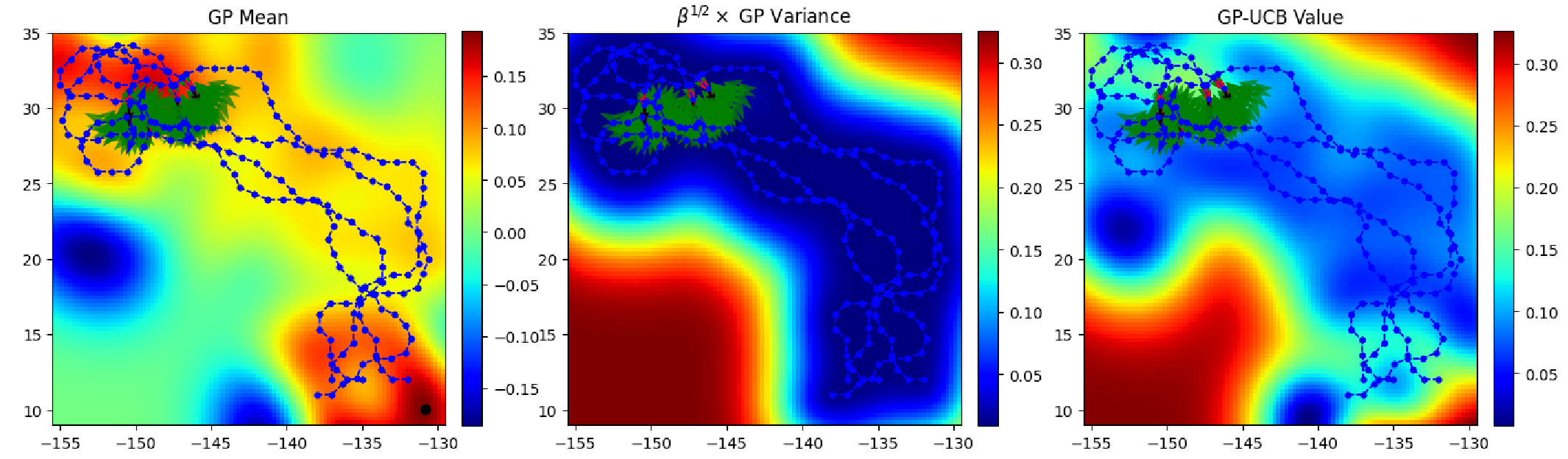}}
    \subfigure[Voronoi partitioning path after the robots have spent 50 units of time.]{\includegraphics[width=0.9\columnwidth]{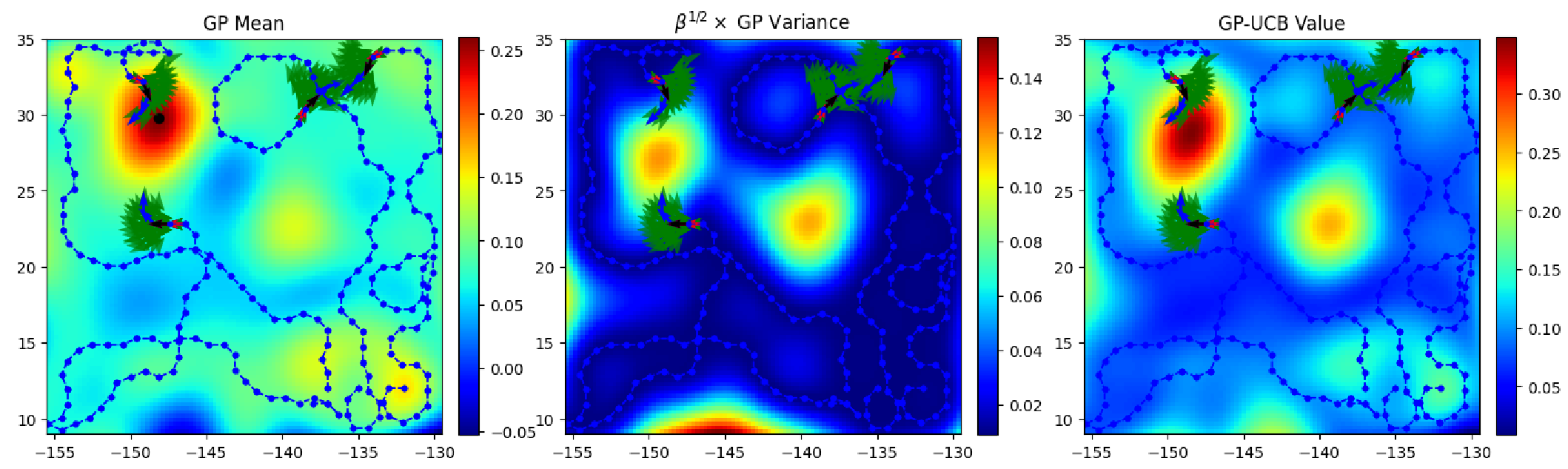}}
	\caption{Voronoi partitioning vs No partitioning comparison at time 50.}
   \label{fig:budget4r50}
\end{figure}

\begin{figure}[htp!]
	\centering
	\includegraphics[width=0.8\columnwidth]{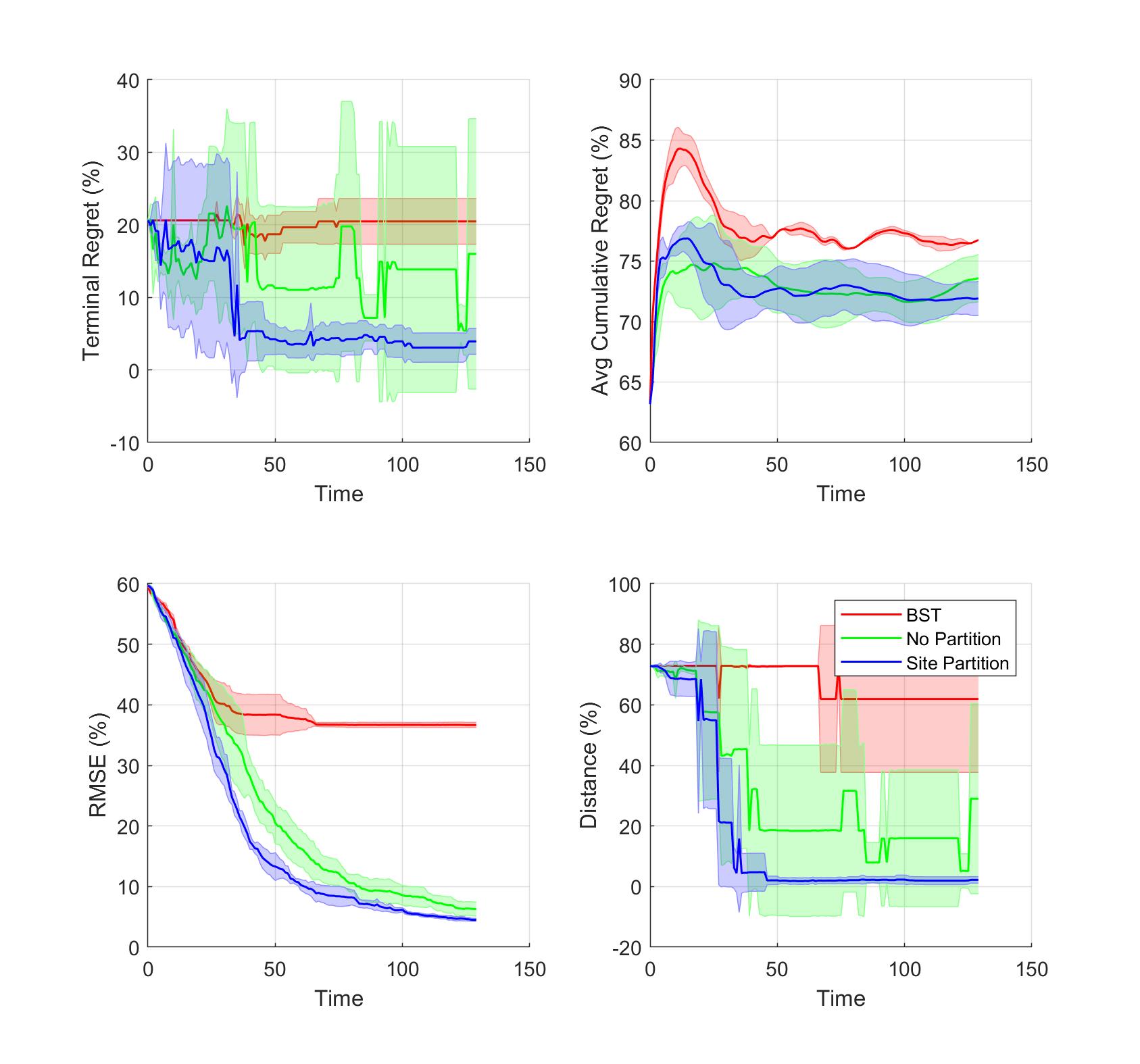}
	\caption {The sensor noise standard deviation was set to 5\%.}
   \label{fig:4rstd5}
\end{figure}

We run ten missions for four robots that start with starting locations (-135, 12), (-132, 12), (-137, 12), (-138, 11)~\ref{fig:env_chlorophyll}. The selected locations near the local maxima divert robots from the hotspot, encouraging them to explore more broadly. Comparing scenarios with and without partitioning shows that partitioning facilitates more uniform exploration and lowers GP variance, as seen in Figure~\ref{fig:budget4r50} after 50 time units. Without it, robots often cover the same areas, leading to redundant measurements. Table~\ref{ch4table:chloromulti} presents metrics analogous to Table~\ref{table:fouthotspot} and Figure~\ref{fig:4rstd5} mirrors Figure~\ref{fig:1rstd5}. The two Voronoi-based methods outperform the Boustrophedon pattern (represented by the red plot). Utilizing Voronoi partitioning offers a distinct edge over not using it (Green plot). Without partitioning, robots risk redundant measurements in overlapping areas. Voronoi partitioning efficiently distributes exploration among robots.

\begin{table}
\centering
\caption{\textnormal{The time budget is 130 units.}}
\begin{tabular}{|c|c|c|}
\hline
BST  &No Partition  &Voronoi\\
 \hline
$20.20\pm 1.91$  &$14.10\pm9.43$  &$7.34\pm2.81$\\
$77.71\pm 0.52$  &$72.81\pm1.60$  &$72.78\pm1.59$\\
$40.14\pm 0.86$  &$22.20\pm1.42$  &$18.21\pm0.67$ \\
$67.54\pm 9.49$  &$30.81\pm 13.71$  &$16.57\pm1.68$\\
\hline
\end{tabular}
\label{ch4table:chloromulti}
\end{table}

\section{Conclusion} \label{sec:conclusion}
In this study, we investigated hotspot identification using both single and multiple mobile sensors. We introduced the AdaptGP-MCTS algorithm designed for scenarios with unknown GP hyperparameters. While adaptive hyperparameter optimization is computationally intensive, our results indicate that AdaptGP-MCTS is a viable option for users with ample mission time but limited computing power on their robotic system. For shorter missions, when the kernel's prior hyperparameters are unknown, OptGP-MCTS emerges as the recommended choice. For multiple robots, we employed a Voronoi region-based partitioning system to delegate specific environment subregions to individual robots. Path planning within these partitions harnesses the UCB values from the learned GP model, enabling robots to strike a balance between exploration and exploitation. As a direction for future research, it would be intriguing to assess the performance of an AdaptGP-MCTS-style planner in multi-robot scenarios.



\bibliographystyle{IEEEtran}
\bibliography{IEEEabrv, IEEEexample}

\begin{thebibliography}{10}
\providecommand{\url}[1]{#1}
\csname url@rmstyle\endcsname
\providecommand{\newblock}{\relax}
\providecommand{\bibinfo}[2]{#2}
\providecommand\BIBentrySTDinterwordspacing{\spaceskip=0pt\relax}
\providecommand\BIBentryALTinterwordstretchfactor{4}
\providecommand\BIBentryALTinterwordspacing{\spaceskip=\fontdimen2\font plus
\BIBentryALTinterwordstretchfactor\fontdimen3\font minus
  \fontdimen4\font\relax}
\providecommand\BIBforeignlanguage[2]{{%
\expandafter\ifx\csname l@#1\endcsname\relax
\typeout{** WARNING: IEEEtran.bst: No hyphenation pattern has been}%
\typeout{** loaded for the language `#1'. Using the pattern for}%
\typeout{** the default language instead.}%
\else
\language=\csname l@#1\endcsname
\fi
#2}}

\bibitem{tokekar2016sensor}
P.~Tokekar, J.~{Vander Hook}, D.~Mulla, and V.~Isler, ``Sensor planning for a
  symbiotic {UAV} and {UGV} system for precision agriculture,'' \emph{{IEEE}
  Transactions on Robotics}, vol.~32, no.~6, pp. 1498--1511, 2016.

\bibitem{suryan2020learning}
V.~Suryan and P.~Tokekar, ``Learning a spatial field in minimum time with a
  team of robots,'' \emph{IEEE Transactions on Robotics (TRO)}, vol.~36, no.~5,
  pp. 1562--1576, Oct. 2020.

\bibitem{aktar2009impact}
W.~Aktar, D.~Sengupta, and A.~Chowdhury, ``Impact of pesticides use in
  agriculture: their benefits and hazards,'' \emph{Interdisciplinary
  toxicology}, vol.~2, no.~1, pp. 1--12, 2009.

\bibitem{blanchard2020informative}
A.~Blanchard and T.~Sapsis, ``Informative path planning for anomaly detection
  in environment exploration and monitoring,'' \emph{arXiv preprint
  arXiv:2005.10040}, 2020.

\bibitem{sung2018online}
Y.~Sung, D.~Dixit, and P.~Tokekar, ``Online multi-robot exploration of a
  translating plume: Competitive algorithm and experiments,'' \emph{arXiv
  preprint arXiv:1811.02769}, 2018.

\bibitem{moore1985usa}
T.~Moore, ``In the usa, robotics technology used at the tmi-2 cleanup and at
  other nuclear plants has prompted interest and shaped research on how robots
  might best be used,'' \emph{IAEA BULLETIN}, p.~31, 1985.

\bibitem{Browne2012}
C.~B. Browne, E.~Powley, D.~Whitehouse, S.~M. Lucas, P.~I. Cowling,
  P.~Rohlfshagen, S.~Tavener, D.~Perez, S.~Samothrakis, and S.~Colton, ``A
  survey of monte carlo tree search methods,'' \emph{IEEE Transactions on
  Computational Intelligence and AI in games}, vol.~4, no.~1, pp. 1--43, 2012.

\bibitem{srinivas2009gaussian}
N.~Srinivas, A.~Krause, S.~M. Kakade, and M.~Seeger, ``Gaussian process
  optimization in the bandit setting: No regret and experimental design,''
  \emph{arXiv preprint arXiv:0912.3995}, 2009.

\bibitem{tanemura1983new}
M.~Tanemura, T.~Ogawa, and N.~Ogita, ``A new algorithm for three-dimensional
  voronoi tessellation,'' \emph{Journal of Computational Physics}, vol.~51,
  no.~2, pp. 191--207, 1983.

\bibitem{inproceedingsKobilarov}
Y.~Teck~Tan, A.~Kunapareddy, and M.~Kobilarov, ``Gaussian process adaptive
  sampling using the cross-entropy method for environmental sensing and
  monitoring,'' in \emph{International Conference on Robotics and Automation
  2018}, 05 2018, pp. 6220--6227.

\bibitem{7526683}
C.~{Mellucci}, P.~P. {Menon}, C.~{Edwards}, and P.~{Challenor}, ``Source
  seeking using a single autonomous vehicle,'' in \emph{2016 American Control
  Conference (ACC)}, 2016, pp. 6441--6446.

\bibitem{DBLP:journals/corr/abs-1809-10611}
\BIBentryALTinterwordspacing
E.~Rolf, D.~Fridovich{-}Keil, M.~Simchowitz, B.~Recht, and C.~J. Tomlin, ``A
  successive-elimination approach to adaptive robotic sensing,'' \emph{CoRR},
  vol. abs/1809.10611, 2018. [Online]. Available:
  \url{http://arxiv.org/abs/1809.10611}
\BIBentrySTDinterwordspacing

\bibitem{Liu-RSS-19}
W.~Chen and L.~Liu, ``Pareto monte carlo tree search for multi-objective
  informative planning,'' in \emph{Proceedings of Robotics: Science and
  Systems}, FreiburgimBreisgau, Germany, June 2019.

\bibitem{10.5555/1625275.1625631}
A.~Singh, A.~Krause, C.~Guestrin, W.~Kaiser, and M.~Batalin, ``Efficient
  planning of informative paths for multiple robots,'' in \emph{Proceedings of
  the 20th International Joint Conference on Artifical Intelligence}, ser.
  IJCAI'07.\hskip 1em plus 0.5em minus 0.4em\relax San Francisco, CA, USA:
  Morgan Kaufmann Publishers Inc., 2007, p. 2204–2211.

\bibitem{krause2008near}
A.~Krause, A.~Singh, and C.~Guestrin, ``Near-optimal sensor placements in
  gaussian processes: Theory, efficient algorithms and empirical studies,''
  \emph{Journal of Machine Learning Research}, vol.~9, no. Feb, pp. 235--284,
  2008.

\bibitem{binney2013optimizing}
J.~Binney, A.~Krause, and G.~S. Sukhatme, ``Optimizing waypoints for monitoring
  spatiotemporal phenomena,'' \emph{The International Journal of Robotics
  Research}, vol.~32, no.~8, pp. 873--888, 2013.

\bibitem{suryan2018sensor}
V.~Suryan and P.~Tokekar, ``Learning a spatial field with gaussian process
  regression in minimum time,'' in \emph{Algorithmic Foundations of Robotics
  XIII}.\hskip 1em plus 0.5em minus 0.4em\relax Cham: Springer International
  Publishing, 2020, pp. 301--317.

\bibitem{kemna2018pilot}
S.~Kemna, O.~Kroemer, and G.~S. Sukhatme, ``Pilot surveys for adaptive
  informative sampling,'' in \emph{2018 IEEE International Conference on
  Robotics and Automation (ICRA)}.\hskip 1em plus 0.5em minus 0.4em\relax IEEE,
  2018, pp. 6417--6424.

\bibitem{xiao2022nonmyopic}
C.~Xiao and J.~Wachs, ``Nonmyopic informative path planning based on global
  kriging variance minimization,'' \emph{IEEE Robotics and Automation Letters},
  2022.

\bibitem{gelly2012grand}
S.~Gelly, L.~Kocsis, M.~Schoenauer, M.~Sebag, D.~Silver, C.~Szepesv{\'a}ri, and
  O.~Teytaud, ``The grand challenge of computer go: Monte carlo tree search and
  extensions,'' \emph{Communications of the ACM}, vol.~55, no.~3, pp. 106--113,
  2012.

\bibitem{browne2012survey}
C.~B. Browne, E.~Powley, D.~Whitehouse, S.~M. Lucas, P.~I. Cowling,
  P.~Rohlfshagen, S.~Tavener, D.~Perez, S.~Samothrakis, and S.~Colton, ``A
  survey of monte carlo tree search methods,'' \emph{IEEE Transactions on
  Computational Intelligence and AI in games}, vol.~4, no.~1, pp. 1--43, 2012.

\bibitem{10.5555/3020751.3020809}
R.~Marchant, F.~Ramos, and S.~Sanner, ``Sequential bayesian optimisation for
  spatial-temporal monitoring,'' in \emph{Proceedings of the Thirtieth
  Conference on Uncertainty in Artificial Intelligence}, ser. UAI'14.\hskip 1em
  plus 0.5em minus 0.4em\relax Arlington, Virginia, USA: AUAI Press, 2014, p.
  553–562.

\bibitem{tolpin2012mcts}
D.~Tolpin and S.~Shimony, ``Mcts based on simple regret,'' in \emph{Proceedings
  of the AAAI Conference on Artificial Intelligence}, vol.~26, no.~1, 2012, pp.
  570--576.

\bibitem{barrientos2011aerial}
A.~Barrientos, J.~Colorado, J.~d. Cerro, A.~Martinez, C.~Rossi, D.~Sanz, and
  J.~Valente, ``Aerial remote sensing in agriculture: A practical approach to
  area coverage and path planning for fleets of mini aerial robots,''
  \emph{Journal of Field Robotics}, vol.~28, no.~5, pp. 667--689, 2011.

\bibitem{kazmi2011adaptive}
W.~Kazmi, M.~Bisgaard, F.~Garcia-Ruiz, K.~D. Hansen, and A.~la~Cour-Harbo,
  ``Adaptive surveying and early treatment of crops with a team of autonomous
  vehicles,'' in \emph{Proceedings of the 5th European Conference on Mobile
  Robots ECMR 2011}, 2011, pp. 253--258.

\bibitem{dhariwal2004bacterium}
A.~Dhariwal, G.~S. Sukhatme, and A.~A. Requicha, ``Bacterium-inspired robots
  for environmental monitoring,'' in \emph{IEEE International Conference on
  Robotics and Automation, 2004. Proceedings. ICRA'04. 2004}, vol.~2.\hskip 1em
  plus 0.5em minus 0.4em\relax IEEE, 2004, pp. 1436--1443.

\bibitem{dunbabin2012robots}
M.~Dunbabin and L.~Marques, ``Robots for environmental monitoring: Significant
  advancements and applications,'' \emph{IEEE Robotics \& Automation Magazine},
  vol.~19, no.~1, pp. 24--39, 2012.

\bibitem{ouimet2013collective}
M.~Ouimet and J.~Cort{\'e}s, ``Collective estimation of ocean nonlinear
  internal waves using robotic underwater drifters,'' \emph{IEEE Access},
  vol.~1, pp. 418--427, 2013.

\bibitem{https://doi.org/10.48550/arxiv.2203.02865}
\BIBentryALTinterwordspacing
G.~P. Kontoudis and D.~J. Stilwell, ``Fully decentralized, scalable gaussian
  processes for multi-agent federated learning,'' 2022. [Online]. Available:
  \url{https://arxiv.org/abs/2203.02865}
\BIBentrySTDinterwordspacing

\bibitem{1013690}
R.~Zlot, A.~Stentz, M.~Dias, and S.~Thayer, ``Multi-robot exploration
  controlled by a market economy,'' in \emph{Proceedings 2002 IEEE
  International Conference on Robotics and Automation (Cat. No.02CH37292)},
  vol.~3, 2002, pp. 3016--3023 vol.3.

\bibitem{doi:10.1177/0278364913515309}
\BIBentryALTinterwordspacing
C.~Nieto-Granda, I.~John G.~Rogers, and H.~I. Christensen, ``Coordination
  strategies for multi-robot exploration and mapping,'' \emph{The International
  Journal of Robotics Research}, vol.~33, no.~4, pp. 519--533, 2014. [Online].
  Available: \url{https://doi.org/10.1177/0278364913515309}
\BIBentrySTDinterwordspacing

\bibitem{5650551}
J.~Yuan, Y.~Huang, T.~Tao, and F.~Sun, ``A cooperative approach for multi-robot
  area exploration,'' in \emph{2010 IEEE/RSJ International Conference on
  Intelligent Robots and Systems}, 2010, pp. 1390--1395.

\bibitem{6385730}
D.~E. Soltero, M.~Schwager, and D.~Rus, ``Generating informative paths for
  persistent sensing in unknown environments,'' in \emph{2012 IEEE/RSJ
  International Conference on Intelligent Robots and Systems}, 2012, pp.
  2172--2179.

\bibitem{6798666}
A.~Marino, G.~Antonelli, A.~P. Aguiar, A.~Pascoal, and S.~Chiaverini, ``A
  decentralized strategy for multirobot sampling/patrolling: Theory and
  experiments,'' \emph{IEEE Transactions on Control Systems Technology},
  vol.~23, no.~1, pp. 313--322, 2015.

\bibitem{7989245}
S.~Kemna, J.~G. Rogers, C.~Nieto-Granda, S.~Young, and G.~S. Sukhatme,
  ``Multi-robot coordination through dynamic voronoi partitioning for
  informative adaptive sampling in communication-constrained environments,'' in
  \emph{2017 IEEE International Conference on Robotics and Automation (ICRA)},
  2017, pp. 2124--2130.

\bibitem{8460473}
W.~Luo and K.~Sycara, ``Adaptive sampling and online learning in multi-robot
  sensor coverage with mixture of gaussian processes,'' in \emph{2018 IEEE
  International Conference on Robotics and Automation (ICRA)}, 2018, pp.
  6359--6364.

\bibitem{Berkenkamp2019}
F.~Berkenkamp, A.~P. Schoellig, and A.~Krause, ``No-regret bayesian
  optimization with unknown hyperparameters,'' \emph{arXiv preprint
  arXiv:1901.03357}, 2019.

\end{thebibliography}

\end{document}